\documentclass[10pt,twocolumn,letterpaper]{article}

\usepackage{iccv}
\usepackage{times}
\usepackage{epsfig}
\usepackage{graphicx}
\usepackage{amsmath}
\usepackage{amssymb}
\usepackage{bm}
\usepackage{url}
\usepackage{breqn}
\usepackage[inline]{enumitem}
\usepackage{wrapfig}
\usepackage{caption}
\usepackage{graphicx}
\usepackage{float}
\usepackage{tabularx}
\usepackage{booktabs}
\usepackage[tight]{subfigure}
\usepackage{multirow}
\usepackage{enumitem}
\setitemize{noitemsep,topsep=0pt,parsep=0pt,partopsep=0pt}
\usepackage{subfiles}
\usepackage{svg}
\usepackage{xfrac}

\newcommand{\Acronym}[0]{NeRFuser\xspace} 

\usepackage[]{xcolor}

\newif\iffinal

\iffinal
    \newcommand{\mw}[1]{}
    \newcommand{\todo}[1]{}
\else
    \newcommand{\mw}[1]{\textcolor{red}{[MW: #1]}}
    \newcommand{\todo}[1]{\textcolor{red}{[TODO: #1]}}
\fi

\definecolor{mygray}{RGB}{89, 89, 89}
\definecolor{mypink}{RGB}{228, 155, 157}
\definecolor{myteal}{RGB}{0, 151, 167}
\definecolor{myblue}{RGB}{66, 133, 244}

\usepackage[bookmarks=true,colorlinks=true,citecolor=blue,linkcolor=blue]{hyperref}

\usepackage{flushend}

\iccvfinalcopy %

\def\iccvPaperID{11244} %

\ificcvfinal\fi

\title{\Acronym: Large-Scale Scene Representation by NeRF Fusion}
\author{Jiading Fang$^{*,1}$, Shengjie Lin$^{*,1}$, Igor Vasiljevic$^2$, Vitor Guizilini$^2$, \\ Rares Ambrus$^2$, Adrien Gaidon$^2$, Gregory Shakhnarovich$^1$, Matthew R. Walter$^1$
}

\begin{document}

\twocolumn[{%
\maketitle

\vspace{-10mm}
\begin{center}
    \centering
    \captionsetup{type=figure}
    \includesvg[height=5.5cm]{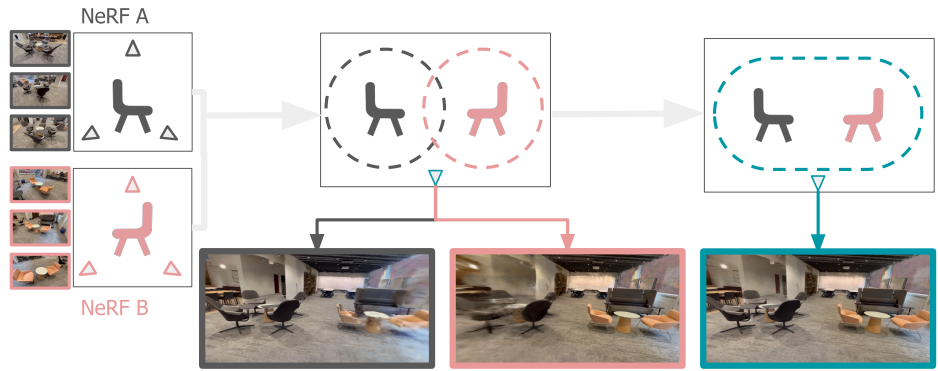}
  \captionof{figure}{
        \textbf{\Acronym.} Starting from separately constructed input NeRFs, \textcolor{mygray}{NeRF $A$} and \textcolor{mypink}
        {NeRF $B$}, we render images at \textcolor{teal}{novel viewpoints}, including those not well-covered by either input NeRF. Our proposed \textit{registration through re-rendering} followed by a novel blending procedure leads to a higher quality result (lower right) than renders from the original NeRFs (\textcolor{mygray}{$A$} or \textcolor{mypink}{$B$}).}
    \label{fig:teaser}
\end{center}
    }]

\maketitle
\ificcvfinal\thispagestyle{empty}\fi

\let\thefootnote\relax\footnotetext{\hspace{-1.3em}Code available at \url{https://github.com/ripl/nerfuser}\\%
$^*$Equal contribution.\\$^1$Toyota Technological Institute at Chicago, Chicago, IL.\\%
$^2$Toyota Research Institute, Los Altos, CA.}

\begin{abstract}
A practical benefit of implicit visual representations like Neural Radiance Fields (NeRFs) is their memory efficiency: large scenes can be efficiently stored and shared as small neural nets instead of collections of images. However, operating on these implicit visual data structures requires extending classical image-based vision techniques (e.g., registration, blending) from image sets to neural fields. Towards this goal, we propose \Acronym, a novel architecture for NeRF registration and blending that assumes only access to pre-generated NeRFs, and not the potentially large sets of images used to generate them.
We propose \textit{registration from re-rendering}, a technique to infer the transformation between NeRFs based on images synthesized from individual NeRFs.
For blending, we propose \textit{sample-based inverse distance weighting} to blend visual information at the ray-sample level. 
We evaluate \Acronym on public benchmarks and a self-collected object-centric indoor dataset, showing the robustness of our method, including to views that are challenging to render from the individual source NeRFs.

\end{abstract}
\vspace{-5mm} 

\section{Introduction}

The transmission of visual data relies on efficient video and image encoding and decoding, technologies with

decades of development behind them.
Most computer vision methods and tools are specialized to this type of data---methods that align and blend images are ubiquitous~\cite{szeliski2007image} and fundamentally designed to work on explicit 2D representations of images.

However, a new learning-based representation for visual data has emerged in recent years: \textit{neural fields}~\cite{neuralfields}. Pioneered by neural radiance fields (NeRFs)~\cite{Mildenhall2020NeRFRS}, these implicit representations allow for efficient visual compression and impressive view synthesis, generating a potentially infinite set of possible views from a fixed set of training images.  Despite the promise of this representation as a storage and communication format, there is a lack of tools that \textit{treat NeRFs as data}, much like common image processing tools treat images.

Towards expanding the utility of NeRFs as a data representation, we propose \Acronym (Fig.~\ref{fig:teaser}), a NeRF fusion framework for the registration and blending of pre-trained NeRFs. Treating input NeRFs as black boxes (i.e., raw data), without access to the images that generate them, our method can register NeRFs (in both pose and scale) and render images from blended NeRFs.  Removing the need for source images also greatly reduces memory consumption. A typical scene may be captured by $100$ images, each about $1$\,MB in size. In contrast, NeRF, which acts as a compression of the individual images, provides an implicit representation of the scene that takes up approximately $5$\,MB, a $20\times$ reduction from the set of original images. Directly transferring this implicit representation makes it possible to build real-time 3D capturing applications (e.g. NeRF streaming).

\begin{figure*}[!ht]
    \centering
    \includegraphics[width=\linewidth]{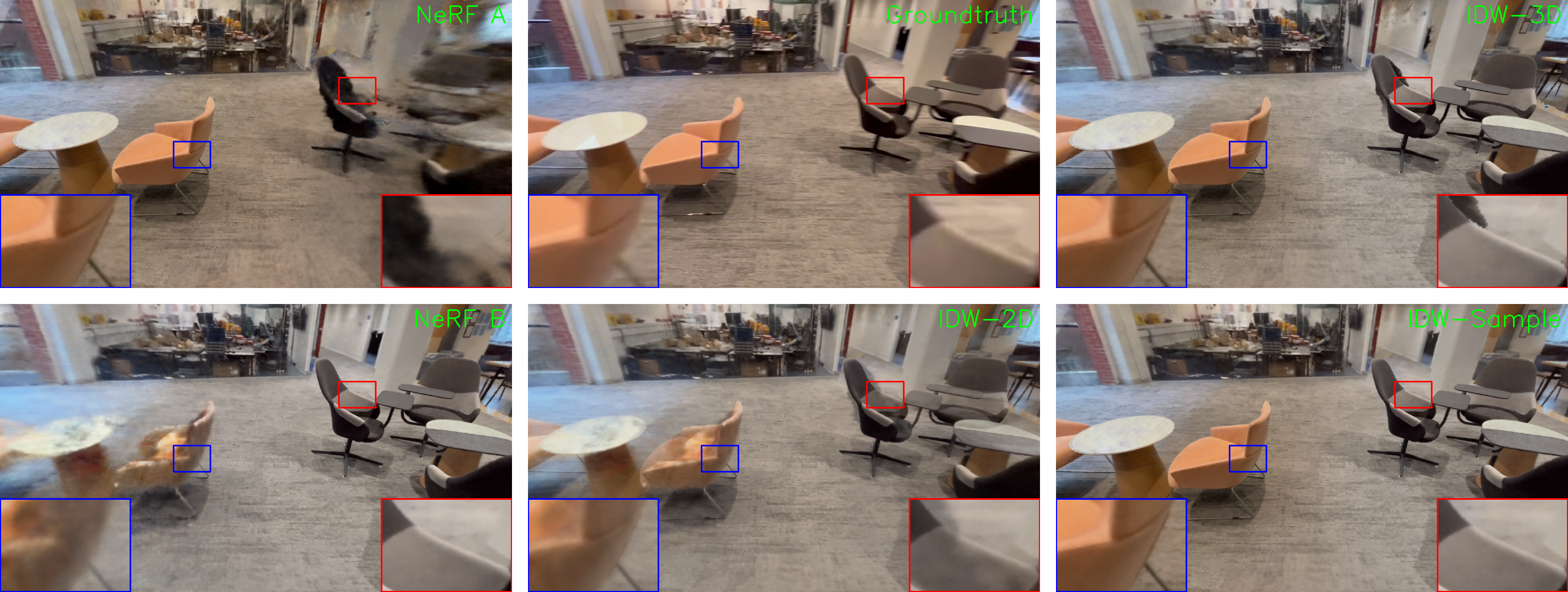}%
    \caption{Qualitative comparison of blending methods. Our proposed \emph{IDW-Sample} produces high-quality blending for both chairs, while baseline methods fail on at least one chair. Notice that the blended results (e.g., \emph{IDW-Sample}) are even sharper than the real test image, which exhibits motion blur, demonstrating an advantage of fusing information from multiple NeRFs.}%
    \label{fig:blending_demo}
\end{figure*}

\Acronym fuses NeRFs in two steps: \textit{registration} and \textit{blending}. For the first step, we propose registration from re-rendering. It takes advantage of the ability of modern NeRFs to synthesize high-quality views, which enables us to make use of a 2D image matching pipeline for the 3D NeRF registration task.
For the second step, inspired by BlockNeRF~\cite{Tancik2022BlockNeRFSL}, we propose a fine-grained sample-based blending method, including a novel compatible weighting method.
In summary, we propose
\begin{enumerate*}[label=(\roman*)]
    \item a \textbf{novel registration from re-rendering method}, that aligns uncalibrated implicit representations and solves for relative scale and pose; and
    \item a \textbf{new blending method to composite predictions from multiple NeRFs}, resulting in blended images that are better than images rendered by any individual NeRF.
\end{enumerate*}
\section{Related Work}
\subsection{Neural Radiance Fields}
A Neural Radiance Field (NeRF)~\cite{Tancik2022BlockNeRFSL} is a parametric representation of a 3D scene. It optimizes a neural network composed of MLPs to encode the scene as density and radiance fields, which can be used to synthesize novel views through volumetric rendering. Since its introduction, many follow-up works~\cite{Barron2021MipNeRFAM,Mller2022InstantNG, Barron2021MipNeRF3U, Yu2021PlenoxelsRF, Sun2021DirectVG, Tancik2023NerfstudioAM} have improved over the original implementation.

One line of improvement involves the reconstruction of large-scale NeRFs~\cite{Xiangli2021CityNeRFBN, Zhang2022NeRFusionFR, Tancik2022BlockNeRFSL, Sucar2021iMAPIM, zhu2022nice, Zhu2023NICERSLAMNI, Turki2021MegaNeRFSC, Reiser2021KiloNeRFSU, Tancik2022BlockNeRFSL}. However, most of these works focus on reconstructing the entire scene with a single model. While progressive training~\cite{Xiangli2021CityNeRFBN, Sucar2021iMAPIM} and carefully designed data structures~\cite{Zhang2022NeRFusionFR, zhu2022nice, Zhu2023NICERSLAMNI} have helped to expand the expressivity of a single model, other works~\cite{Reiser2021KiloNeRFSU, Tancik2022BlockNeRFSL} have shown that a collection of many small models can perform better, while maintaining the same number of parameters. Our method provides a novel way to reason over many small models, combining them to improve performance.

\subsection{NeRF Registration}

NeRFs are optimized from posed images, with poses usually obtained using a structure-from-motion (SfM) method \cite{schoenberger2016SfM, schoenberger2016mvs, sarlin2019coarse, sarlin2020superglue, DeTone2017SuperPointSI, Revaud2019R2D2RA, Dusmanu2019D2NetAT}. Because these methods are scale-agnostic, the resulting coordinate system will have an arbitrary scale specific to each NeRF. Jointly using multiple NeRFs requires \emph{NeRF Registration}, i.e.,solving for the relative transformation between their coordinate systems.

Note that the setting is different from ``NeRF Inversion''~\cite{yen2021inerf, lin2022parallel} that estimates the 6-DoF camera pose relative to the pre-trained NeRF given an image, a technique that has been used for NeRF-based visual navigation and localization~\cite{Adamkiewicz2021VisionOnlyRN, Maggio2022LocNeRFMC, Chen2023CATNIPSCA}. However, these tasks can potentially be handled by \Acronym if formulated as NeRF-to-NeRF pose estimation. Also relevant are works that jointly optimize NeRF representations along with the poses and intrinsics~\cite{lin2021barf, wang2021nerf, jeong2021self}. However, \Acronym only uses SfM on re-rendered images, and does not modify the pre-trained NeRFs themselves.

While there is a large body of work on registration for explicit representations (e.g., point-clouds)~\cite{Pomerleau2015ARO}, there are few works on NeRF registration. To the best of our knowledge, there are two works related to ours: nerf2nerf~\cite{Goli2022nerf2nerfPR} and Zero-NeRF~\cite{Peat2022ZeroNR}. Both approaches perform registration in a purely geometric way---extracting surfaces from the NeRF representations, and thus do not take full advantage of the rich encoded radiance information and its rendering capability. Further, nerf2nerf is only capable of \emph{local} registration under a known scale, and even then requires a reasonable initialization in the form of human annotations. On the other hand, our method performs scaled \emph{global} registration and does not require any human annotations.

\subsection{NeRF Blending}
Image blending is a highly researched topic in computational photography~\cite{Burt1983TheLP, Brown2003RecognisingP}, but few works have discussed blending in terms of NeRFs. Relevant is Block-NeRF~\cite{Tancik2022BlockNeRFSL}, which blends NeRFs in both image- and pixel-wise manner. It introduces two ways to measure the blending weights of NeRFs: inverse distance weighting (IDW) and visibility prediction.
IDW computes the contribution of each NeRF according to:
\begin{equation}\label{eq:idw}
    w_i \propto d_i^{-\gamma},
\end{equation}
where $d_i$ is some notion of distance between the camera and elements of NeRF $i$, $\gamma\in\mathcal{R}$ is a positive hyper-parameter that modulates the blending rate.

Building on this work, we propose a blending approach that operates on ray samples, as well as a new IDW method for sample-wise blending. \Acronym provides a principally more refined way of blending, which we show leads to sharper images.

\section{Methodology} \label{sec:methodology}
In this section we will first describe our NeRF registration method \textit{registration from re-rendering} and then our blending technique \textit{IDW-Sample}.
\subsection{Background}\label{sec:background}
In this section we include background information on volumetric rendering and inverse distance weighing (IDW)~\cite{Tancik2022BlockNeRFSL} methods for neural radiance field (NeRF)~\cite{Mildenhall2020NeRFRS} blending.

\subsubsection{Volumetric Rendering}\label{sec: nerf-rendering}
Given a 3D point $\bm{p}$ and a viewing direction $\bm{d}$, NeRF $\mathcal{R}$ predicts the scene density $\sigma$ at that point and its color $\bm{c}$ when viewed along direction $\bm{d}$ 
\begin{equation}
    [\sigma(\bm{p}),\bm{c}(\bm{p},\bm{d})]=\mathcal{R}(\bm{p},\bm{d}).
\end{equation}
To render novel views, consider the image pixel corresponding to camera ray $\bm{r}=(\bm{o},\bm{d})$, where $\bm{o}$ is the camera's optical center and $\bm{d}$ is the ray direction.

In practice, non-overlapping samples are proposed along the ray at locations with significant predicted density. Assuming $K$ intervals of length $\delta_k$ sampled along ray $\bm{r}$ at a distance of $t_k$ from $\bm{o}$, a NeRF predicts the density and color for each sample as
\begin{subequations}
    \begin{align}
        \sigma_k &= \sigma(\bm{o}+t_k\bm{d}),\\
        \bm{c}_k &= \bm{c}(\bm{o}+t_k\bm{d},\bm{d}).
    \end{align}
\end{subequations}
The accumulated color is
\begin{equation}\label{eq:vr}
    \bm{C}(\bm{r})=\sum_{k=1}^nT_k\alpha_k\bm{c}_k,
\end{equation}
where 
$\alpha_k=1-\exp(-\sigma_k\delta_k)$ 
is the probability that light is blocked in this sampled interval of length $\delta_k$ along the ray at location $t_k$. The probability of light reaching this interval (i.e., not being blocked along the way) is then
\begin{equation}
    T_k =\prod_{l=1}^{k-1}(1-\alpha_l)
        = \exp\left(-\sum_{l=1}^{k-1}\sigma_l\delta_l\right).
\end{equation}
Additionally, termination probability is defined as
\begin{equation}
    p_k=T_k\alpha_k,
\end{equation}
which is the probability of light traveling along $\bm{r}$ and getting blocked at sample $k$. Equation~\ref{eq:vr} thus becomes

\begin{equation}\label{eq:vr2}
    \bm{C}(\bm{r})=\sum_{k=1}^np_k\bm{c}_k.
\end{equation}

\subsubsection{NeRF Blending with IDW}
\label{sec:block-nerf-blending}
When using multiple NeRFs to render an image, the contribution of each NeRF can be determined using inverse distance weighting (IDW) 
\begin{equation}\label{eq:idw}
    w_i \propto d_i^{-\gamma},
\end{equation}
where $d_i$ is the Euclidean distance of the blending element (e.g. ray sample in \emph{IDW-Sample}) to NeRF $i$'s origin, $\gamma\in\mathbb{R}^+$ is a hyper-parameter that modulates the blending rate.
Block-NeRF~\cite{Tancik2022BlockNeRFSL} proposes two variants of IDW, namely \emph{IDW-2D} and \emph{IDW-3D}, which we discuss below. 

\vspace{2ex}
\noindent\textbf{IDW-2D} blends images using an image-wise weighting
\begin{equation}
    I = \sum_i w_i I_i,
\end{equation}
where we use the distance between the query camera center and NeRF $i$'s origin as $d_i$ to compute $w_i$. While \emph{IDW-2D} works well when the query camera is much closer to one of the NeRFs than the others, it suffers when the query camera is roughly of the same distance from all source NeRFs. In the later case, the blended image will be a blurry mixture affected by noisy regions existing in source images, resulting in poor visual quality.

\vspace{2ex}
\noindent\textbf{IDW-3D} is a pixel-wise weighting strategy that considers the distance between the origin $\bm{x}_i$ of each NeRF $i$ and the 3D coordinates $\bm{p}_i^{(j)}$ of pixel $j$ determined using the expected depth predicted by NeRF $i$,
\begin{equation}
    d_i^{(j)} = \lVert \bm{x}_i - \bm{p}_i^{(j)}\rVert_2.
\end{equation}
Each pixel $j$ is then rendered by substituting $d_i^{(j)}$ into Equation~\ref{eq:idw} as
\begin{equation}
    I^{(j)} = \sum_i w_i^{(j)}I_i^{(j)}.
\end{equation}
The major problem with \emph{IDW-3D} is that, to accurately obtain the expected point of ray termination $\bm{p}_i^{(j)}$, it requires NeRF $i$ to faithfully predict the depth for pixel $j$. This is not always fulfilled since
\begin{enumerate*}[label=(\roman*)]
    \item NeRFs are not known to accurately reconstruct the scene geometry; and moreover,
    \item source NeRFs will be focusing on different portions of the scene by design, leading to invalid blending weights.
\end{enumerate*}
Empirically, \emph{IDW-3D} usually performs the worst among all blending methods.

\subsection{Registration from Re-Rendering}
The first step of our pipeline is to estimate the relative transformations between two or more input NeRFs.

We assume that each NeRF is trained on a separate set of images, capturing different, yet overlapping, portions of the same scene (i.e., each input NeRF has at least one neighbor).
We do not assume a specific type of training data, e.g., the poses used to train the NeRFs may have come from KinectFusion~\cite{Newcombe2011KinectFusionRD} and have metric scale, or they may be the result of a SfM pipeline (e.g., COLMAP~\cite{schoenberger2016mvs}) and thus have arbitrary scale.

As a result, each NeRF may have a unique coordinate system that is inconsistent with the others, in terms of translation and rotation as well as scale. Without loss of generality, the following discussion focuses on the registration of two NeRFs, $A$ and $B$, as the extension to more than two NeRFs is straightforward.

Our goal is to find the transformation $T_{BA}\in\mathrm{SIM}(3)$ that transforms a 3D point $p_B$ in NeRF $B$ to its corresponding point $p_A$ in NeRF $A$ as $p_A = T_{BA} p_B$.
Note that $T_{BA} = \left[\begin{smallmatrix} R_{BA} & \bm{t}_{BA}\\
        \bm{0} & 1 \end{smallmatrix} \right]S_{BA}$ can be decomposed into a rotation $R_{BA}$, translation $\bm{t}_{BA}$, and uniform scaling of factor $s_{BA}$,
where $S_{BA}$ is a diagonal matrix $\mathrm{diag}(s_{BA}, s_{BA}, s_{BA}, 1)$.

First, we assume that the NeRFs are produced with sufficient training views, so that they can generate high quality novel views. We then sample a set of poses (e.g. uniformly on the upper hemisphere), which we use as local poses to query both NeRFs to get re-renderings.

We re-purpose off-the-shelf structure-from-motion methods (SuperPoint as the feature~\cite{DeTone2017SuperPointSI} and SuperGlue as the matcher~\cite{sarlin2020superglue}) on the union of re-rendered images from the two NeRFs in order to recover their poses in the same coordinate system, which we then use for registration, as discussed next.

\noindent\textbf{Procedure and Notation}
~Given the trained model of NeRF $A$, we query it with sampled camera poses $\left\{G_{A_i}^A\right\}_i$ to synthesize images $\left\{I_{A_i}\right\}_i$. $G_{A_i}^A\in\mathrm{SE}(3)$ is specified as a pose matrix that transforms a point from the coordinate system of camera $A_i$ to that of NeRF $A$. $I_{A_i}$ is the image synthesized from NeRF $A$ using the query pose $G_{A_i}^A$. Likewise, we query NeRF $B$ with $\left\{G_{B_i}^B\right\}_i$ to synthesize images $\left\{I_{B_i}\right\}_i$. We then feed images $\left\{I_{A_i}\right\}_i\cup\left\{I_{B_i}\right\}_i$ as input to SfM, and obtain poses $\left\{G_{A_i}^C\right\}_i$ and $\left\{G_{B_i}^C\right\}_i$ as output. $G_{A_i}^C\in\mathrm{SE}(3)$ is the recovered pose of image $I_{A_i}$ from SfM. It is specified as a pose matrix that transforms a point from the coordinate frame of camera $A_i$ to $C$, where $C$ is the coordinate system determined by this SfM execution. Note that an $\mathrm{SE}(3)$ pose matrix does not involve scale, so the induced camera-to-world transformation always assumes a specific camera instance that shares the same scale as the world.

\noindent\textbf{Recovering Scale}~
Let $S_{AC}=\mathrm{diag}(s_{AC}, s_{AC}, s_{AC}, 1)$ be the scale matrix from NeRF $A$ to $C$, meaning that one unit length in NeRF A equals $s_A$ units in $C$. Considering $G_{ij}\in\mathrm{SE}(3)$ as the pose of camera $A_i$ relative to camera $A_j$ when specified in $C$'s scale, we have
\begin{equation}\label{eq:1}
    \begin{aligned}
        G_{ij}  &={G_{A_j}^C}^{-1}G_{A_i}^C&&\text{using \textit{C} as bridge}\\
                &=S_{AC}{G_{A_j}^A}^{-1}G_{A_i}^A{S_{AC}}^{-1}&&\text{using \textit{A} as bridge}
    \end{aligned}
\end{equation}
If we further dissect $G_{A_i}^C$ as
$
    \begin{bmatrix}
        R_{A_i}^C & \bm{t}_{A_i}^C\\
        \bm{0} & 1
    \end{bmatrix}
$
and repeat this for $G_{A_j}^C,G_{A_i}^A,G_{A_j}^A$, Equation~\ref{eq:1} becomes
\begin{dmath}\label{eq:2}
    \begin{bmatrix}
        {R_{A_j}^C}^\top R_{A_i}^C & {R_{A_j}^C}^\top (\bm{t}_{A_i}^C-\bm{t}_{A_j}^C)\\
        \bm{0} & 1
    \end{bmatrix}=\begin{bmatrix}
        {R_{A_j}^A}^\top R_{A_i}^A & s_{AC}{R_{A_j}^A}^\top (\bm{t}_{A_i}^A-\bm{t}_{A_j}^A)\\
        \bm{0} & 1
    \end{bmatrix}.
\end{dmath}
Note that all components involved in Equation~\ref{eq:2} are either determined when sampling camera poses or given by SfM with the exception of $s_{AC}$, which is what we want to recover. Specifically, equating the L2 norm of the translation part from both sides of Equation~\ref{eq:2} gives
\begin{equation}\label{eq:3}
    s_{AC}=\frac{\lVert \bm{t}_{A_i}^C-\bm{t}_{A_j}^C\rVert_2}{\lVert \bm{t}_{A_i}^A-\bm{t}_{A_j}^A\rVert_2}
\end{equation}
In practice, we use the median over all $i,j$ pairs to construct $S_{AC}$. $S_{BC}$ is recovered similarly.

\noindent\textbf{Recovering Transformations}~
Let $T_{AC}\in\mathrm{SIM}(3)$ be the transformation from NeRF $A$ to $C$. Using camera $A_i$ as bridge, we have $T_{AC}=G_{A_i}^CS_{AC}{G_{A_i}^A}^{-1}$.
In practice, we compute $T_{AC}$ over all instances of camera $A_i$, and choose the closest valid $\mathrm{SIM}(3)$ transformation to the median result. $T_{BC}$ is recovered similarly. We then compute NeRF $B$ to NeRF $A$ transformation as $T_{BA}={T_{AC}}^{-1}T_{BC}$.

\noindent\textbf{Robustness to pose estimation errors}~
\label{sec:registration_robustness}
While our proposed registration method works better with more accurate SfM results on NeRF-synthesized images, it is also robust to SfM's errors. When computing the relative scale, we only need to recover at least two poses (so that at least one pair is formed to be used in Equation~\ref{eq:3}) from each NeRF's re-renderings. This is easily achievable with a reasonably sampled set of query poses. Moreover, since we consider the median result as the final estimation, the impact of erroneous poses will be minimal. A similar analysis also holds for transformation recovery, except that only a single pose is needed for the estimation.

\subsection{NeRF Blending}
Given two or more registered NeRFs and a query camera pose, NeRF blending~\cite{Tancik2022BlockNeRFSL} aims to combine predictions from the individual NeRFs with the goal of high-quality novel view synthesis. Without loss of generality, we consider again the two-NeRF setting: $A$ and $B$ with relative transformation $T_{BA}\in\mathrm{SIM}(3)$. Let $G_B\in\mathrm{SE}(3)$ be a pose defined in NeRF $B$'s coordinate system that can be used to query NeRF $B$. To get the corresponding pose $G_A$ to query NeRF $A$, we first decompose $T_{BA}=G_{BA}S_{BA}$, and compute $G_A=T_{BA}G_B{S_{BA}}^{-1}$.

For blending, there are three key concepts to consider:
\begin{enumerate*}[label=(\roman*)]
    \item\textbf{when to blend}: in what case should it be used;\label{itm:blend-q1}
    \item\textbf{what to blend}: at what granularity should it happen; and\label{itm:blend-q2}
    \item\textbf{how to blend}: in which way should we compute blending weights?\label{itm:blend-q3}
\end{enumerate*}
Block-NeRF~\cite{Tancik2022BlockNeRFSL} answers \ref{itm:blend-q1} with \textit{visibility thresholding}, where if the mean visibility of a frame (predicted by a visibility network) is above a certain threshold then blending is activated.  Afterwards, it answers \ref{itm:blend-q2} by introducing image- and pixel-wise blending. Finally, it handles \ref{itm:blend-q3} by inverse-distance-weighting (IDW) and predicted visibility weighting. Importantly, to achieve any of these results, a visibility prediction network has to be trained jointly with the NeRF and used during inference. In our setting, we do not assume access to a visibility network, since we are dealing with black-box uncalibrated NeRFs not generated for this particular purpose.

In this paper, we answer \ref{itm:blend-q1} by proposing a simpler threshold that is solely based on distance. We answer \ref{itm:blend-q2} by proposing a novel sample-based blending method recognizing the fact that the color of a pixel is computed using samples along the ray in NeRF during volumetric rendering. We answer \ref{itm:blend-q3} by proposing an IDW method for our sample-based blending. Since we use IDW with sample-based blending, we coin our method \emph{IDW-Sample}.

Without loss of generality, we discuss the blending of two registered NeRFs, $A$ and $B$. Our findings easily extend to an arbitrary number of NeRFs and also to any volumetric representation. 

\subsubsection{Distance Test for NeRF Selection}
The decision of when to render using blended NeRFs, rather than just one NeRF, is an important question, because NeRFs can only render with high-quality within their effective range. Rendering using distant NeRFs, whose rendering quality is poor, can only be harmful. Hence, we introduce a test based on the distance between the origin of the query camera and the NeRF centers. Denoting the distances from NeRF $A$ and NeRF $B$ as $d_A$ and $d_B$, the test value is $\tau = \max{\left(\frac{d_A}{d_B}, \frac{d_B}{d_A}\right)}$.
If $\tau$ is greater than a threshold, it means that the second-closest NeRF is sufficiently far, in which case it is better to simply use the rendering of the closest NeRF to the query camera; otherwise, IDW-based blending is enabled.
\subsubsection{IDW-Sample for NeRF Blending}

During NeRF's volumetric rendering stage, a pixel's color is computed using samples along the ray. Recognizing this fact, we propose a sample-wise blending method that calculates the blending weights for each ray sample using IDW. We show that the original volumetric rendering methodology can be easily extended to take advantage of these new sample-wise blending weights, resulting in our proposed \emph{IDW-Sample} strategy. 
\begin{figure}[!t]
    \centering
    \includesvg[width=\linewidth]{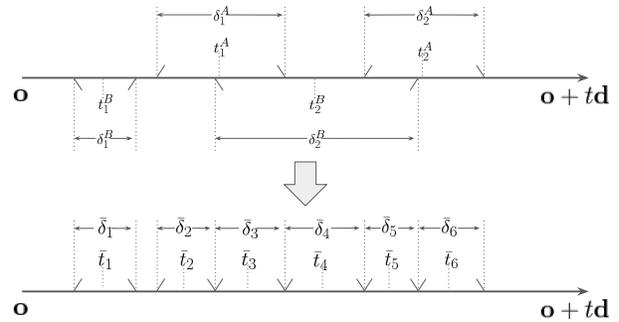}
    \caption{An illustration of how ray samples proposed by two NeRFs are merged based on their locations and lengths. Top: two sets of ray samples proposed by NeRF $A$ and NeRF $B$; Bottom: the single set of merged ray samples.}
    \label{fig:merge_samples}
\end{figure}
\noindent\textbf{Merge Ray Samples}~Consider a pixel to be rendered, which gets unprojected into a ray. Since ray samples are separately proposed according to the density field of each source NeRF, we need to merge them into a single set. As illustrated in Figure~\ref{fig:merge_samples}, given samples $\{(t^A_k,\delta^A_k)\}_k$ and $\{(t^B_k,\delta^B_k)\}_k$ proposed from NeRF $A$ and NeRF $B$, respectively, we merge them into a single set of ray samples $\{(\bar{t}_k, \bar{\delta}_k)\}_k$ by taking the sample location $t$ and length $\delta$. We update the termination probability and color of each new sample in the merged set for each source NeRF. Given a ray sample proposed by a NeRF, we assume its termination probability mass is uniformly distributed over its length, while its color is the same for any point within coverage. 
    
\begin{figure}[!t]
    \centering
    \includesvg[width=\linewidth]{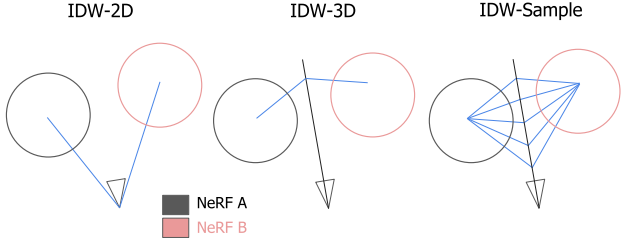}
    \caption{Illustration of IDW-based blending methods. IDW-2D depends on the distance between camera center and NeRF centers. IDW-3D depends on the distance between estimated ray depth and  NeRF centers. IDW-Sample depends on the NeRF centers and sample positions that are irrespective to depth quality, which is major downside of IDW-3D.}
    \label{fig:idw}
\end{figure}
\noindent\textbf{Blending Process}~We use IDW to compute the blending weight for each sample. Specifically, let $\bm{x}_i$ be the origin of $\textrm{NeRF}_i$ for $i\in\{A,B\}$, $\bm{o}$ be the camera's optical center, $\bm{r}=(\bm{o},\bm{d})$ be the ray corresponding to pixel $j$ to be rendered, and $(\bar{t}_k,\bar{\delta}_k)$ be a ray sample from the merged samples set. We compute its blending weight as $w_{i,k}\propto{}{d_{i,k}}^{-\gamma}$, where $d_{i,k}=\|\bm{x}_i-(\bm{o}+\bar{t}_k\bm{d})\|_2$.

The blended pixel $j$ is
\begin{equation}
    I^{(j)}=\sum_k\sum_iw_{i,k}\bar{p}_{i,k} \bar{\bm{c}}_{i,k}
\end{equation}

Weights $w_{i,k}$ are normalized following two steps:
\begin{enumerate*}[label=(\roman*)]
    \item$\sum_iw_{i,k}=1,\;\forall k$; and \label{itm:normalize1}
    \item$\sum_k\sum_iw_{i,k}\bar{p}_{i,k}=1$.\label{itm:normalize2}
\end{enumerate*}
Step~\ref{itm:normalize1} indicates that our method does not change the relative weighting of samples along a given ray, which is already dictated by the termination probability. Step~\ref{itm:normalize2} ensures that the rendered pixel has a valid color. Figure~\ref{fig:idw} provides an illustration.

\begin{table*}[t!]
    \centering
    \begin{tabularx}{0.8\linewidth}{Xcccccc} 
        \toprule
        \multirow{2}{*}{Blending} & \multicolumn{3}{c}{Ground-truth $T_{BA}$} & \multicolumn{3}{c}{Estimated $\hat{T}_{BA}$} \\
        \cmidrule{2-7}
        & PSNR $\uparrow$ & SSIM $\uparrow$ & LPIPS $\downarrow$ & PSNR $\uparrow$ & SSIM $\uparrow$ & LPIPS $\downarrow$ \\ 
        \midrule
        NeRF & 20.92 & 0.716 & 0.369 & 20.90 & 0.714 & 0.370 \\
        Nearest & 23.81 & 0.779 & 0.283 & 23.68 & 0.774 & 0.287 \\ 
        IDW-2D & 24.70 & 0.795 & 0.267 & 24.64 & 0.792 & 0.267 \\
        IDW-3D & 23.48 & 0.776 & 0.279 & 23.45 & 0.772 & 0.280 \\
        IDW-Sample (Ours) & \textbf{24.91} & \textbf{0.813} & \textbf{0.228} & \textbf{24.83} & \textbf{0.810} & \textbf{0.229} \\
        \bottomrule
    \end{tabularx}
    \caption{Blending results on Object-Centric Indoor Scenes. \emph{IDW-Sample} works the best for all metrics with both ground-truth and estimated transformations. Results with estimated $\hat{T}_{BA}$ are only marginally worse than those with ground-truth $T_{BA}$, which demonstrates that our proposed NeRF registration is accurate enough for the downstream blending task.}
    \label{tab:nf_bld}
\end{table*}

\section{Experiments}\label{sec:experiments}
In this section, we describe our registration and blending experiments on Scannet, Block-NeRF, and an object-centric scene dataset we collect, which we will make available.

\subsection{Datasets}
\paragraph{Object-Centric Indoor Scenes}\label{subsec:ttic-data}
We created a dataset consisting of three indoor scenes, using an iPhone $13$ mini in video mode. Each scene consists of three video clips -- we choose two objects in each scene, and collect two (overlapping) video sequences that focus on each object.  Then, we collect a third (test) sequence that observes the entire scene.

We extract images at $3.75$\,fps from all three video clips and feed them jointly to an structure-from-motion (SfM) tool (we use the hloc toolbox~\cite{sarlin2019coarse, sarlin2020superglue}) to recover their poses. These poses are defined in a shared coordinate system, which we denote as $C$. We then center and normalize the poses for each training set, so that the processed poses are located within the bounding box $[-1,1]^3$. Note that this step induces a local coordinate system for each NeRF, which we denote as $A$ and $B$ respectively. We record the resulting transformations $\{T_{CA},T_{CB}\}\subset\mathrm{SIM}(3)$ and treat them as ground-truth. NeRFs $A$ and $B$ are then trained separately from the corresponding training set of images. We test \Acronym on this dataset and report results of both registration and blending in Section~\ref{subsec:nerf-fusion}.

\begin{figure*}[h!]
    \centering
    \begin{tabular}{ccc}
        \centering
        \includegraphics[scale=0.1]{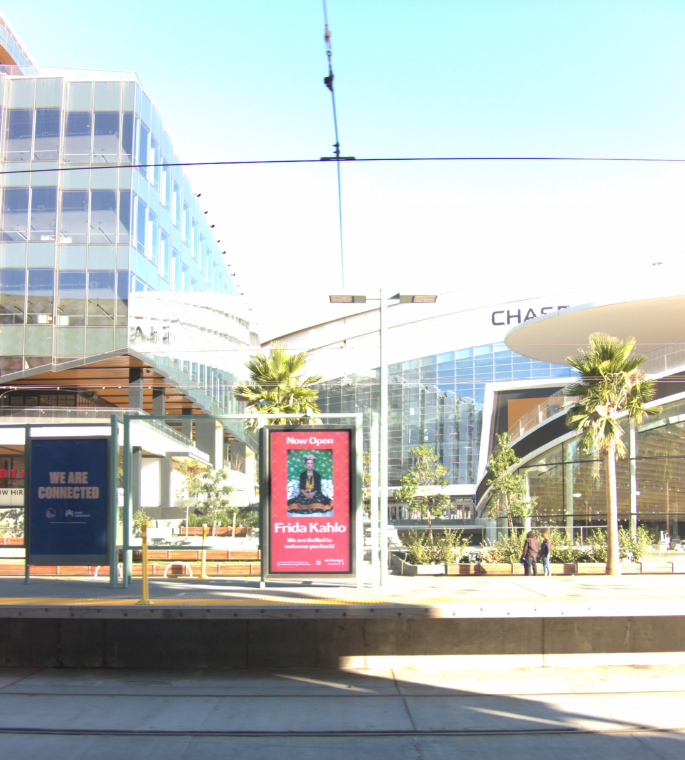} & \includegraphics[scale=0.1]{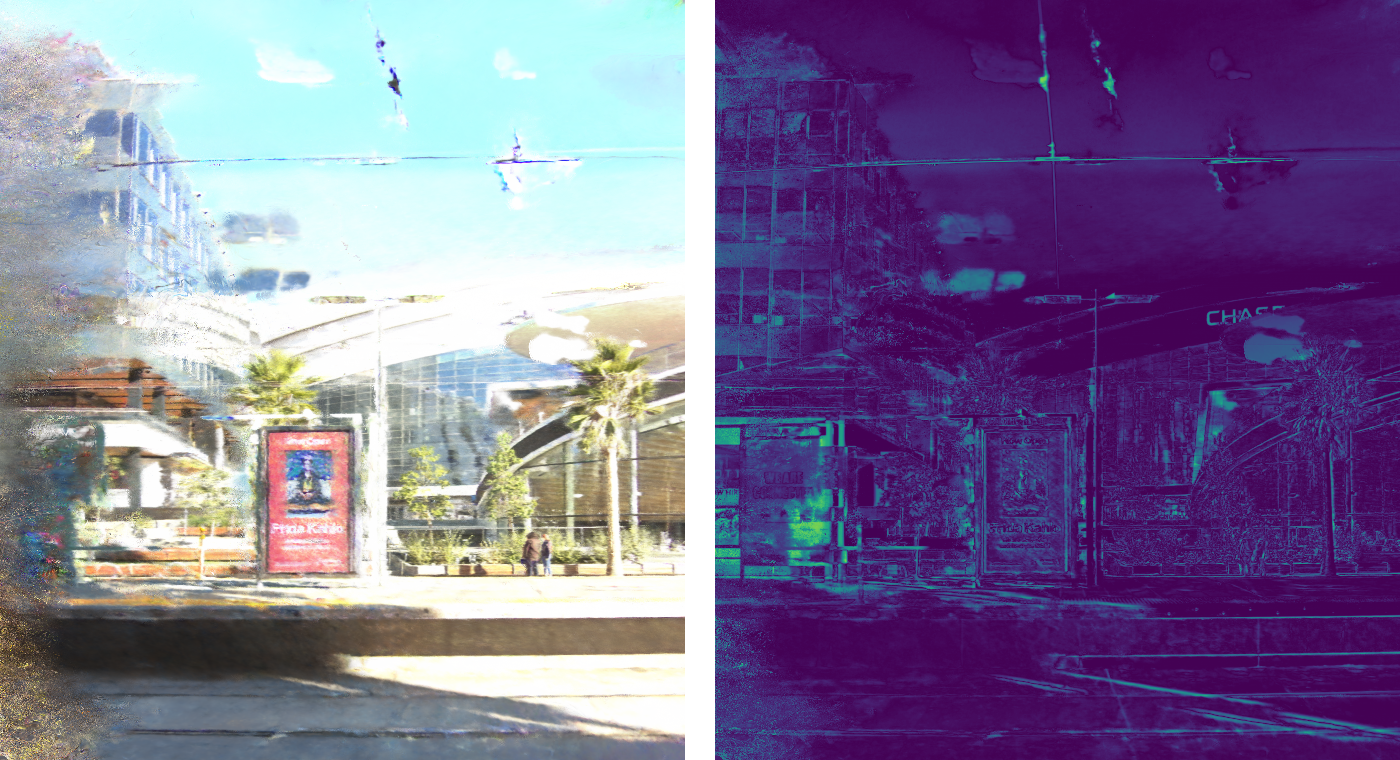} & \includegraphics[scale=0.1]{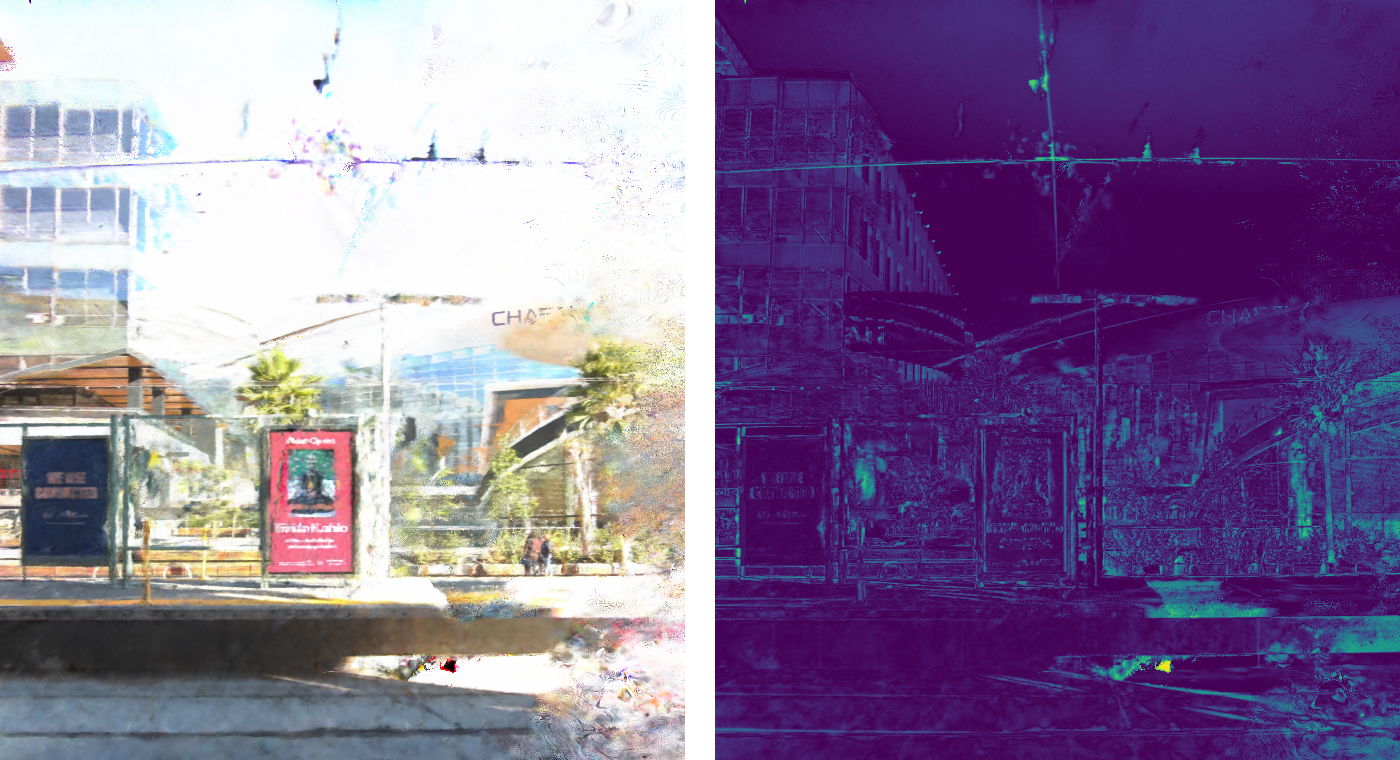}\\
        \cmidrule(lr){1-1}\cmidrule(lr){2-2}\cmidrule(lr){3-3}
        Ground-truth & NeRF A & NeRF B \\
        \includegraphics[scale=0.1]{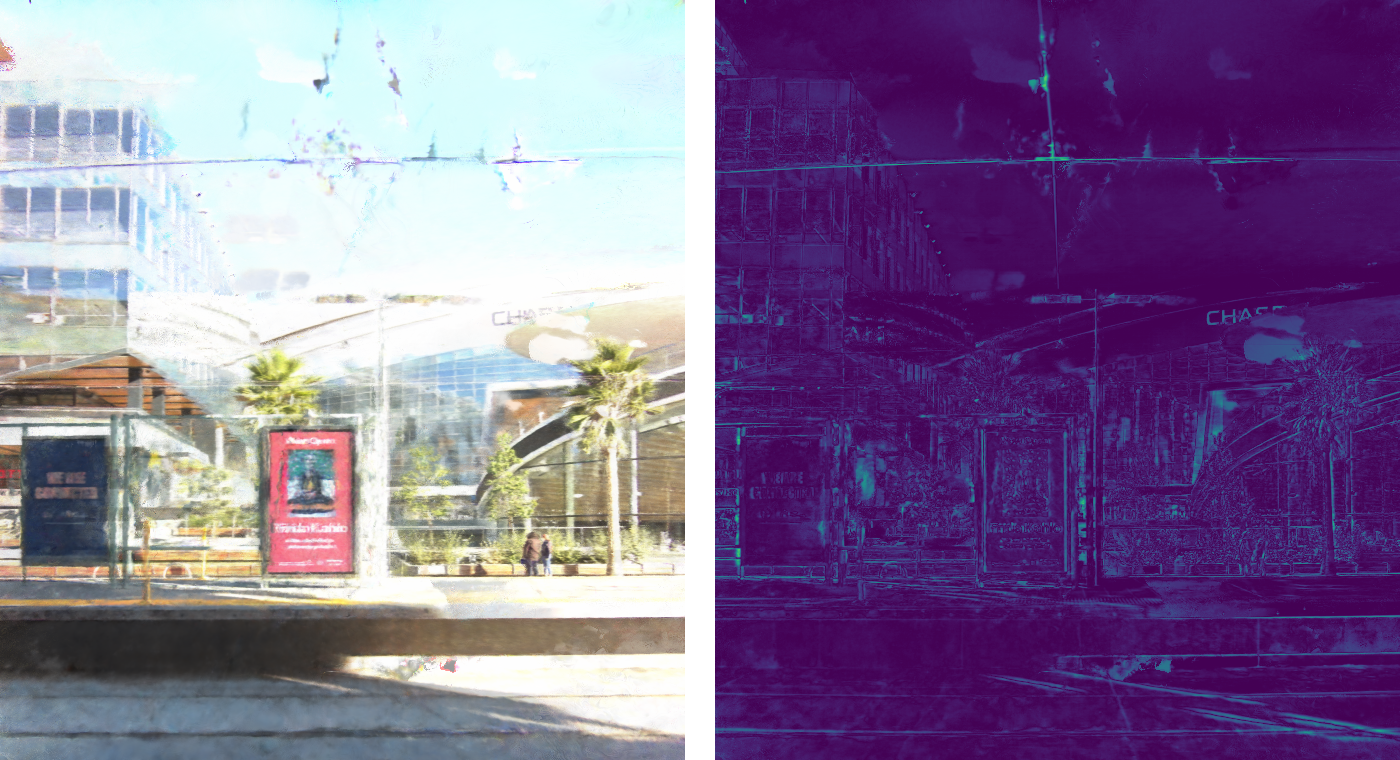} & \includegraphics[scale=0.1]{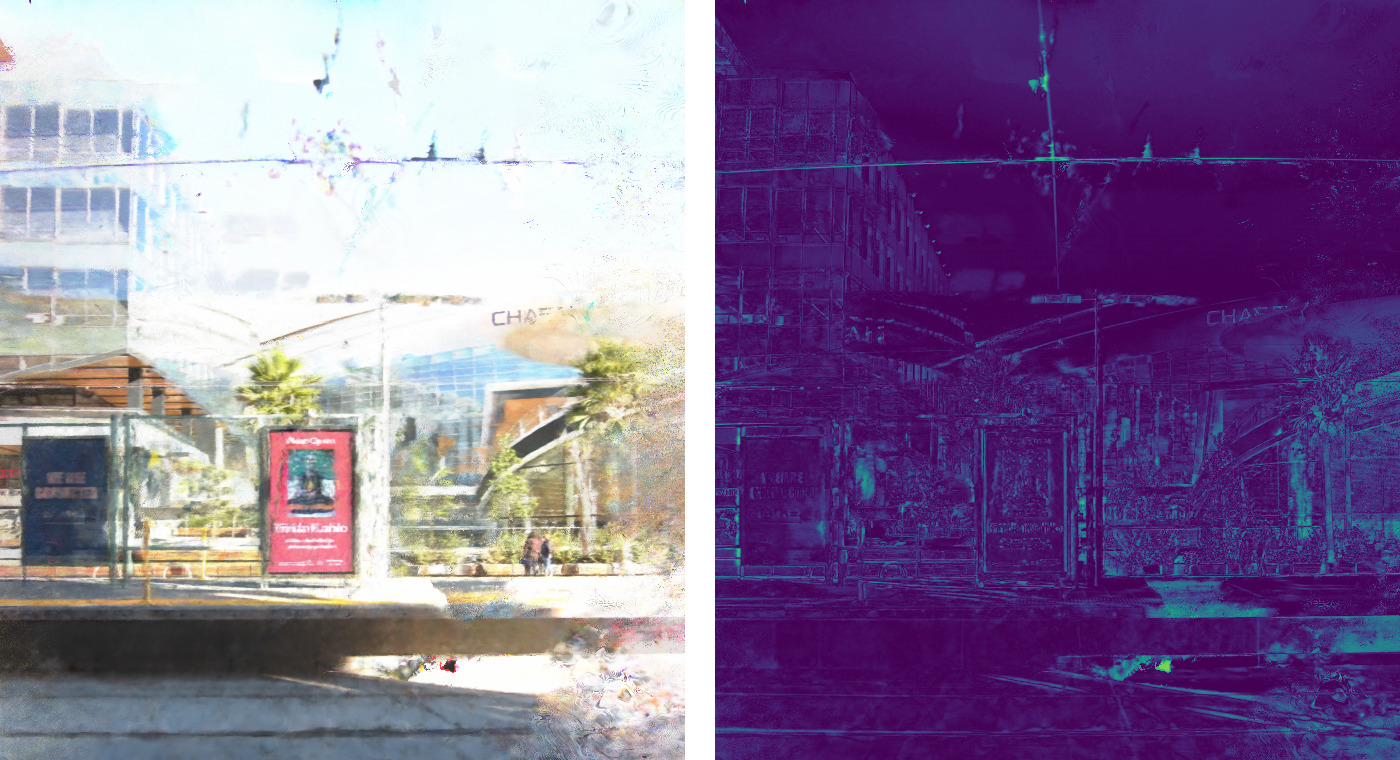} & \includegraphics[scale=0.1]{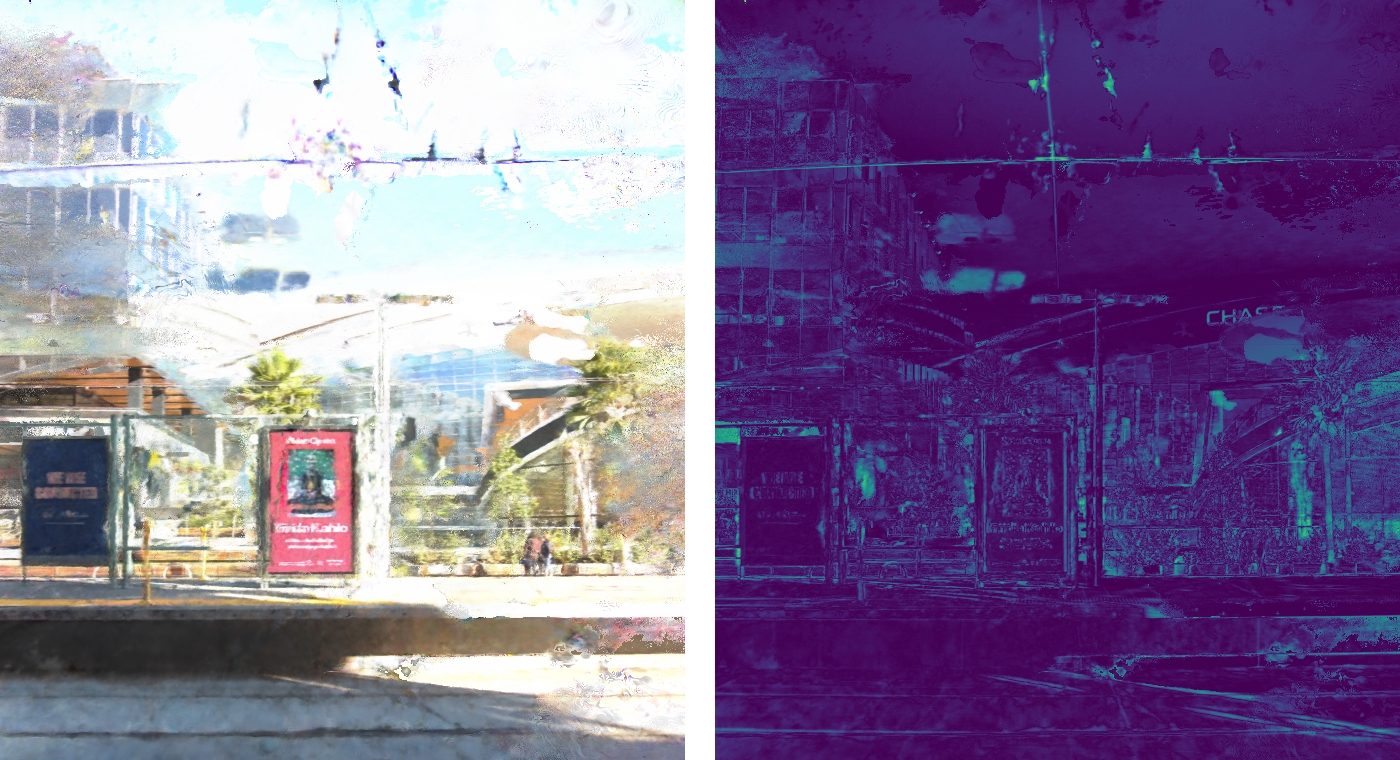}\\
        \cmidrule(lr){1-1}\cmidrule(lr){2-2}\cmidrule(lr){3-3}
        IDW-Sample & IDW-2D & IDW-3D \\
    \end{tabular}%
    \caption{NeRF blending with IDW-based methods on the Mission Bay dataset. Per-pixel errors are visualized as heat maps. Individual NeRFs renderings have large artifacts on either side, which are best resolved by \emph{IDW-Sample} blending.}
    \label{fig:blocknerf-blending}
\end{figure*}

\begin{figure}[h!]
    \centering
    \includesvg[width=\linewidth]{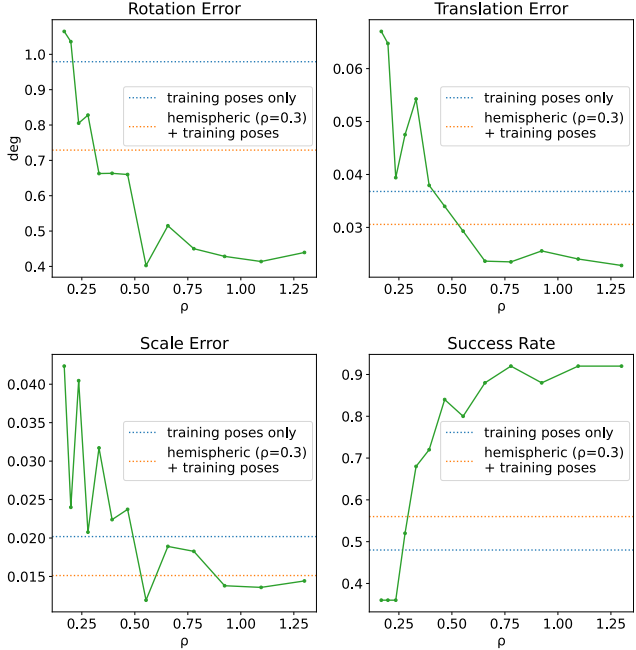}
    \caption{Effect of re-rendering poses on NeRF registration. With more sampled poses, the registration errors go down while success rate improves (\textcolor{teal}{green curve}). Using additional hemispheric poses besides the training ones also proves helpful (\textcolor{orange}{orange line} vs. \textcolor{blue}{blue line}). More interestingly, with a large enough ratio $\rho$, registration with hemispherically sampled poses outperforms training poses when using the same number or fewer poses in total. It shows that it is beneficial to have a larger spatial span of re-rendering poses for registration as illustreated in Figure~\ref{fig:train-vs-rerendering-poses}.}
    \label{fig:ablation-reg-r}
    \vspace{-6mm}
\end{figure}

\paragraph{ScanNet Dataset}
\label{subsubsec:scannet-data}\
Since the ``ground-truth'' poses of our dataset are estimated using SfM based only on RGB images, they are potentially not as accurate as what could be obtained using RGB-D data. Hence, we use the ScanNet dataset to further test registration performance. The ScanNet dataset provides a total of $1513$ RGB-D scenes with annotated camera poses, from which we use the first $218$ scenes. We downsample the frames so that roughly $200$ posed RGB-D images are kept from each scene. We then split the images into three sets: two for training NeRFs and one for testing. Specifically, images are split according to their temporal order. We first randomly select $10\%$ of all images as the test set. Of the remaining images, we label the first $25\%$ as training set $A$, the last $25\%$ as training set $B$, and randomly label the middle $50\%$ as either $A$ or $B$. This splitting strategy creates a moderate spatial overlap among $A$, $B$ and the test sets. Once we have the splits, we center and normalize the training poses the same way as in Section~\ref{subsec:ttic-data}. The resulting transformations $T_{CA},T_{CB}$ are recorded as ground-truth. After generaring the NeRFs, we check their quality according to validation PSNR and keep the best $25$ scenes. We test NeRF registration on this dataset and compare with point-cloud registration in Section~\ref{subsec:ner-reg}.

\paragraph{Mission Bay Dataset}\label{sec:blocknerf-data}
To further test the rendering quality of different blending methods, we run experiments on the Mission Bay dataset from Block-NeRF~\cite{Tancik2022BlockNeRFSL}, which features a street scene from a single capture. The dataset is collected using $12$ cameras that capture the surround of a car that drives along a straight street with a quarter turn at the end. The dataset has approximately $12500$ images, with poses recovered by odometry. We split the images into training and test sets as follows: for every $30$ time-stamped images, we use the first $25$ to construct a block for training and reserve the remaining $5$ for testing. This results in a total of $31$ training sets that we use to construct NeRFs, and $30$ test sets between every $2$ neighboring blocks. We test the blending of two neighboring NeRFs using only the side-view camera (cam73) from the test sets, for which the difference of NeRF renderings are the most visible. We report results in Section~\ref{subsec:nerf-blending}.

\subsection{NeRF Fusion}\label{subsec:nerf-fusion}
We test \Acronym including both NeRF registration and NeRF blending on Object-Centric Indoor Scenes. For registration, we generate $32$ poses that are roughly uniformly placed on the upper hemisphere of radius $1$, with elevation from $0$ to $30^\circ$. We use them as local poses to query NeRFs $A$ and $B$, and feed the $64$ synthesized images jointly to SfM. NeRF $B$ to NeRF $A$ transformation $\hat{T}_{BA}$ is recovered using our proposed procedure. To evaluate its accuracy, given ground-truth and estimated transformations $\{T,\hat{T}\}\subset\mathrm{SIM}(3)$, we first compute $\Delta{}T=\hat{T}T^{-1}$. It is then decomposed into $\Delta{}G\in\mathrm{SE}(3)$ and $\Delta{}S\in\mathrm{SIM}(3)$ as $\Delta{}T=\Delta{}G\Delta{}S$. Rotation error $r_{\textrm{err}}$ is computed as the angle (in degrees) of $\Delta{}G$'s rotation matrix. Translation error $t_{\textrm{err}}$ is computed as the L2 norm of $\Delta{}G$'s translation vector. Note that by definition, $t_{\textrm{err}}$ is measured in NeRF $A$'s unit. For scale error, we extract $\Delta{}s=\lvert\Delta{}S\rvert^{\sfrac{1}{3}}$ and compute $s_{\textrm{err}}=\lvert\log\Delta{}s\rvert$. We report $T_{BA}$ errors against ground-truth: $r_{\textrm{err}}=0.031^\circ,t_{\textrm{err}}=0.0013,s_{\textrm{err}}=0.0045$. 

For blending, we set distance test ratio $\tau=1.8$ and blending rate $\gamma=5$. Since it depends on the NeRF $B$ to NeRF $A$ transformation, we report results in two settings:
\begin{enumerate*}[label=(\roman*)]
    \item using ground-truth $T_{BA}$ and
    \item using estimated $\hat{T}_{BA}$.
\end{enumerate*}
The second setting is specifically used to showcase the compound performance of the full \Acronym framework. In addition to IDW-based methods, we also include \emph{NeRF} and \emph{Nearest} as baselines. \emph{NeRF} directly uses NeRF-synthesized images, while \emph{Nearest} uses the rendering from the closer NeRF to the query pose. We evaluate blending results against ground-truth images on three metrics: PSNR~\cite{Hor2010ImageQM}, SSIM~\cite{Wang2004ImageQA} and LPIPS~\cite{Zhang2018TheUE}. We report numbers averaged over test images of all three scenes from our dataset in Table~\ref{tab:nf_bld}.

\subsection{NeRF Registration}\label{subsec:ner-reg}
To further test the registration performance on a large-scale dataset, we use the ScanNet dataset~\cite{Dai2017ScanNetR3} as prepared according to Section~\ref{subsubsec:scannet-data}. We repeat the same registration procedure as detailed in Section~\ref{subsec:nerf-fusion}, except that $60$ hemispheric poses are sampled instead of $32$. During experiments, we notice failure cases due to NaN or outlier values. To report more meaningful numbers, we treat cases that meet any of the following conditions as failure:
\begin{enumerate*}[label=(\roman*)]
    \item is NaN or
    \item $r_{\textrm{err}}>5^\circ$ or
    \item $t_{\textrm{err}}>0.2$ or
    \item $s_{\textrm{err}}>0.1$.
\end{enumerate*}

We also compare our method against various point-cloud registration (PCR) baselines using both
\begin{enumerate*}[label=(\roman*)]
    \item point-clouds extracted from NeRFs and\label{itm:pcr1}
    \item point-clouds fused from ground-truth posed RGB-D images.\label{itm:pcr2}
\end{enumerate*}
We describe the point-cloud data preparation for each scene as below. While there are scale-adaptive methods~\cite{Sahillioglu2021ScaleAdaptiveI} for PCR, most available and well-tested implementations presume that the two point-clouds to be registered are measured in the same unit. Since $T_{BA}$ is measured in NeRF $A$'s unit, we make sure that the measuring unit of both point-clouds is the same as NeRF $A$'s. For \ref{itm:pcr1} NeRF-extracted point-clouds, the unit conversion is achieved by applying $S_{BA}$ to point-cloud $B$. For \ref{itm:pcr2} RGB-D fused point-clouds, the unit conversion is achieved by applying $S_{CA}$ to both point-clouds. Additionally, note that in this case all ground-truth poses are defined in the same world coordinate system. To enable a fair comparison, we further transform RGB-D fused point-cloud $A$ by $G_{BA}$ after unit conversion. After these processing steps, we get point-clouds ready for registration, whose ground-truth solution is $G_{BA}$ for both \ref{itm:pcr1} and \ref{itm:pcr2}. We report in Table~\ref{table:scannet_registration_filenames_segmentation_nerf_extraction} the results of our registration method and various PCR baselines averaged over all successfully registered scenes, as well as the success rate.

\begin{table}[t!]
    \renewcommand{\arraystretch}{1.0}
    \centering
    \begin{tabularx}{\linewidth}{Xcccc} 
        \toprule
        Registration & $r_{\textrm{err}}~(^\circ)$ & $t_{\textrm{err}}$ & $s_{\textrm{err}}$ & Success\\
        \midrule
        \multicolumn{5}{c}{NeRF-extracted point-cloud} \\
        \midrule[0.1pt]
        ICP \cite{Rusinkiewicz2001EfficientVO} & 3.027 & 0.1151 & N/A & 0.13 \\ 
        FGR \cite{Zhou2016FastGR} & 4.549 & 0.1844 & N/A & 0.04 \\
        FPFH \cite{Rusu2009FastPF} & 2.805 & 0.0381 & N/A & 0.17 \\
        \midrule
        \multicolumn{5}{c}{RGB-D-fused point-cloud} \\
        \midrule[0.1pt]
        ICP \cite{Rusinkiewicz2001EfficientVO} & 1.598 & 0.0816 & N/A & 0.17 \\ 
        FGR \cite{Zhou2016FastGR} & 1.330 & 0.0372 & N/A & 0.71 \\
        FPFH \cite{Rusu2009FastPF} & \textbf{0.049} & \textbf{0.0205} & N/A & \textit{0.79} \\
        \midrule[0.1pt]
        \Acronym & \textit{0.588} & \textit{0.0315} & \textbf{0.0211} & \textbf{0.84}\\
        \bottomrule
    \end{tabularx}
    \caption{\textbf{Registration results on ScanNet.} We compare to point-cloud registration methods on both NeRF-extracted point-cloud and ground-truth RGB-D-fusion point-cloud. Due to the noisy geometry of NeRF reconstructions, registration performance on NeRF-extracted point-clouds is inferior. However, \Acronym is comparable to the registration performance on RGB-D-fused methods in terms of $r_{\textrm{err}}$ and $t_{\textrm{err}}$, while having the highest success rate. Our method also recovers the relative scale, while point-cloud baselines work on the relaxed problem of $\mathrm{SE}(3)$ pose recovery.
    $\textbf{Bold}$ numbers are the best, $\textit{italic}$ numbers are second best.}
    \label{table:scannet_registration_filenames_segmentation_nerf_extraction}
\end{table}

\begin{figure}[!t]
    \centering
    \includesvg[width=0.99\linewidth]{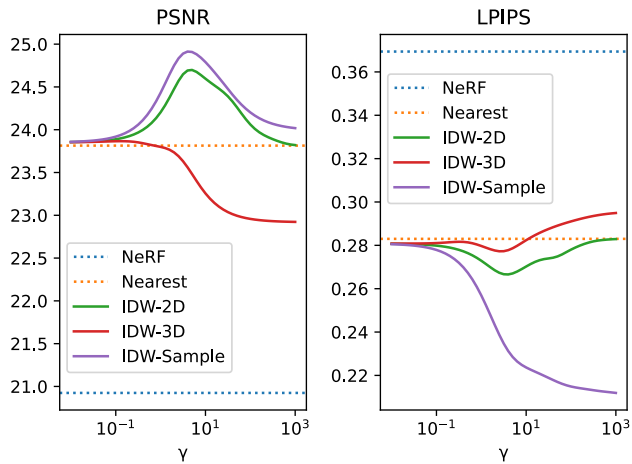}
    \caption{Effect of blending rate $\gamma$ in IDW-based blending ranging from $[0.01, 1000]$. For all blending methods, quality initially increases with $\gamma$, but then decreases as $\gamma$ increases further. }
    \label{fig:gamma-ablation}
\end{figure}

\subsection{NeRF Blending}\label{subsec:nerf-blending}
To further test our blending performance, we use the outdoor Mission Bay dataset as described in Section~\ref{sec:blocknerf-data}, with ground-truth transformations. We set the distance test ratio $\tau=1.2$, and the blending rate $\gamma=10$. Quantitative results averaged over test images of all scenes are reported in Table~\ref{table:blocknerf_blending}. Qualitative results are visualized in Figure~\ref{fig:blocknerf-blending}. 

\subsection{Ablation Studies}
\paragraph{Ablation on re-rendering poses for NeRF registration}
We study the registration performance on ScanNet dataset w.r.t. the number of sampled poses. To account for the fact that each scene may be of a different scale, we introduce $\rho$ as the ratio of the number of sampled poses over the number of training views. We geometrically sample $\rho\in[0.167, 1.3]$, and generate the hemispheric poses accordingly. We evaluate the performance of NeRF registration w.r.t. $\rho$ averaged over all ScanNet scenes. In addition, we include 2 more settings.
\begin{enumerate*}[label=(\roman*)]
    \item\emph{training poses only}: instead of hemispheric poses, use NeRF's training poses for re-rendering;
    \item\emph{hemispheric $+$ training poses}: use NeRF's training poses together with hemispherically sampled ones ($\rho=0.3$) for re-rendering.
\end{enumerate*}
Results are reported in Figure~\ref{fig:ablation-reg-r}. We want to additionally highlight that, since registration using a relatively small number of sampled poses for re-rendering can still work well, it implies that the registration procedure will not take long. In practice, it typically only takes minutes to finish.

\begin{table}[t!]
    \centering
    \renewcommand{\arraystretch}{1.0}
    \begin{tabularx}{\linewidth}{Xccc}
        \toprule
        Blending & PSNR $\uparrow$ & SSIM $\uparrow$ & LPIPS $\downarrow$ \\ 
        \midrule
        NeRF & 17.306 & 0.571 & 0.502 \\
        Nearest & 19.070 & 0.657 & 0.398 \\ 
        IDW-2D & 19.692 & 0.659 & 0.413 \\
        IDW-3D & 18.806 & 0.636 & 0.433 \\
        IDW-Sample (Ours) & \textbf{19.986} & \textbf{0.678} & \textbf{0.388} \\
        \bottomrule
    \end{tabularx}
    \caption{Blending results on Mission Bay dataset. \emph{IDW-Sample} performs the best for all metrics.}
    \label{table:blocknerf_blending}
    \vspace{-3mm}
\end{table}
\begin{figure}[!th]
    \centering
    \includegraphics[width=0.7\linewidth]{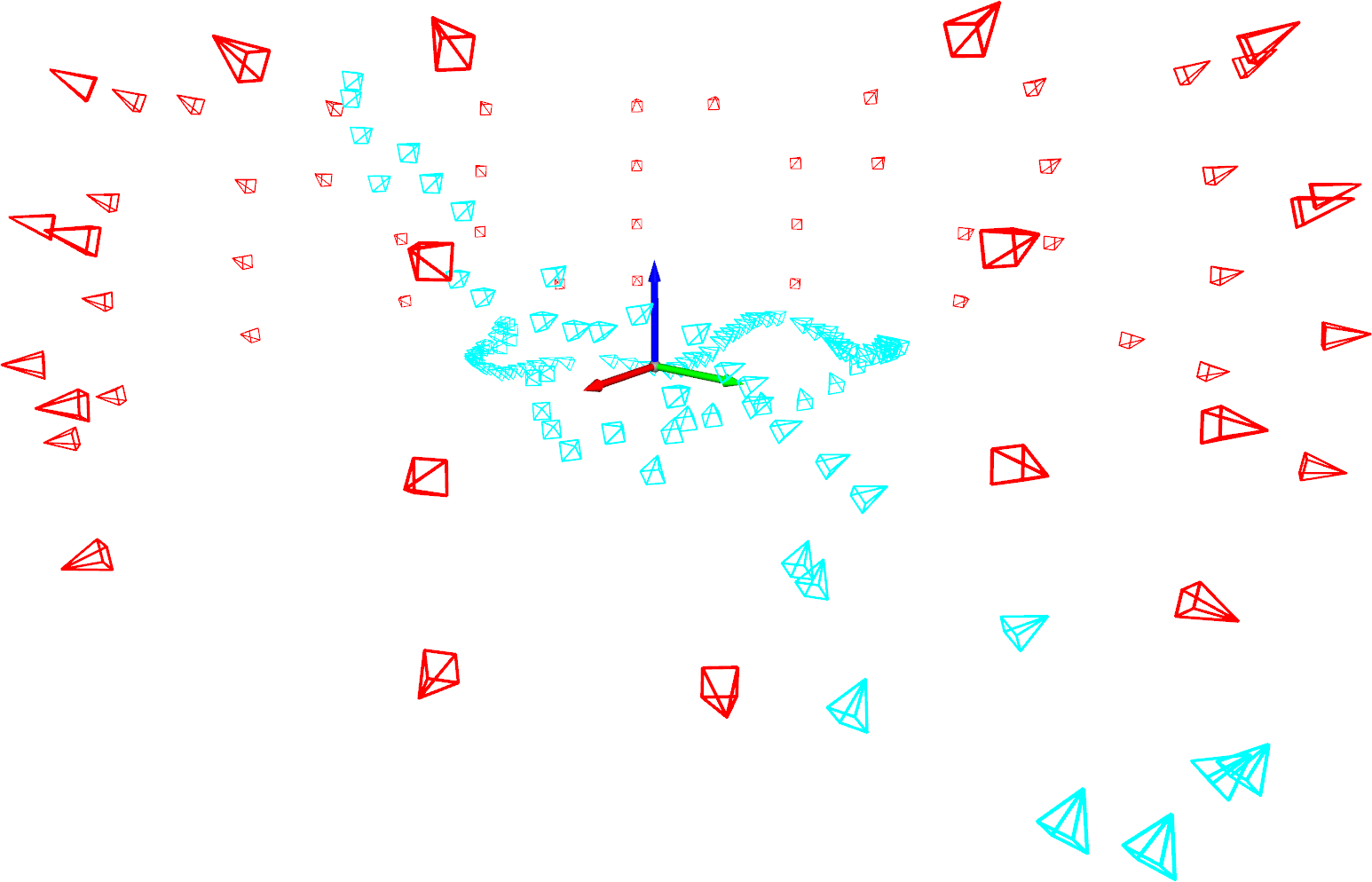}
    \caption{Distributions of \textcolor{cyan}{training poses} and \textcolor{red}{sampled poses} on a ScanNet scene used for NeRF registration. Sample poses are less cluttered than training ones that come from a handheld camera trajectory, which results in a wider baseline for easier SfM.}
    \label{fig:train-vs-rerendering-poses}
\end{figure}

\paragraph{Ablation of $\gamma$ in IDW-based blending}\label{sec:gamma-ablation}
We study the effect of blending rate $\gamma$ in IDW-based blending on Object-Centric Indoor Scenes. Specifically, we use ground-truth transformations and set distance test ratio $\tau=1.8$. We geometrically sample $\gamma$ in $[10^{-2},10^3]$. For each sampled $\gamma$, we blend NeRFs with all IDW-based methods and report the results averaged over test images of all 3 scenes from Object-Centric Indoor Scenes. Since \emph{Nearest} and \emph{NeRF} are not affected by $\gamma$, we draw dotted horizontal lines for comparison. The results are shown in Figure~\ref{fig:gamma-ablation}. In $\gamma\rightarrow{}0$ case, all IDW-based methods become the same as using the mean image. In $\gamma\rightarrow{}\infty$ case, \emph{IDW-2D} becomes the same as \emph{Nearest}, while \emph{IDW-Sample} becomes analogous to KiloNeRF~\cite{Reiser2021KiloNeRFSU} (more details in appendix \ref{subsec:kilo-nerf}). We find the optimal $\gamma$ in between the extremes for any IDW-based method. Moreover, our proposed \emph{IDW-Sample} almost always performs the best for any given $\gamma$.

\begin{figure}[!t]
    \centering
    \includegraphics[width=0.6\linewidth]{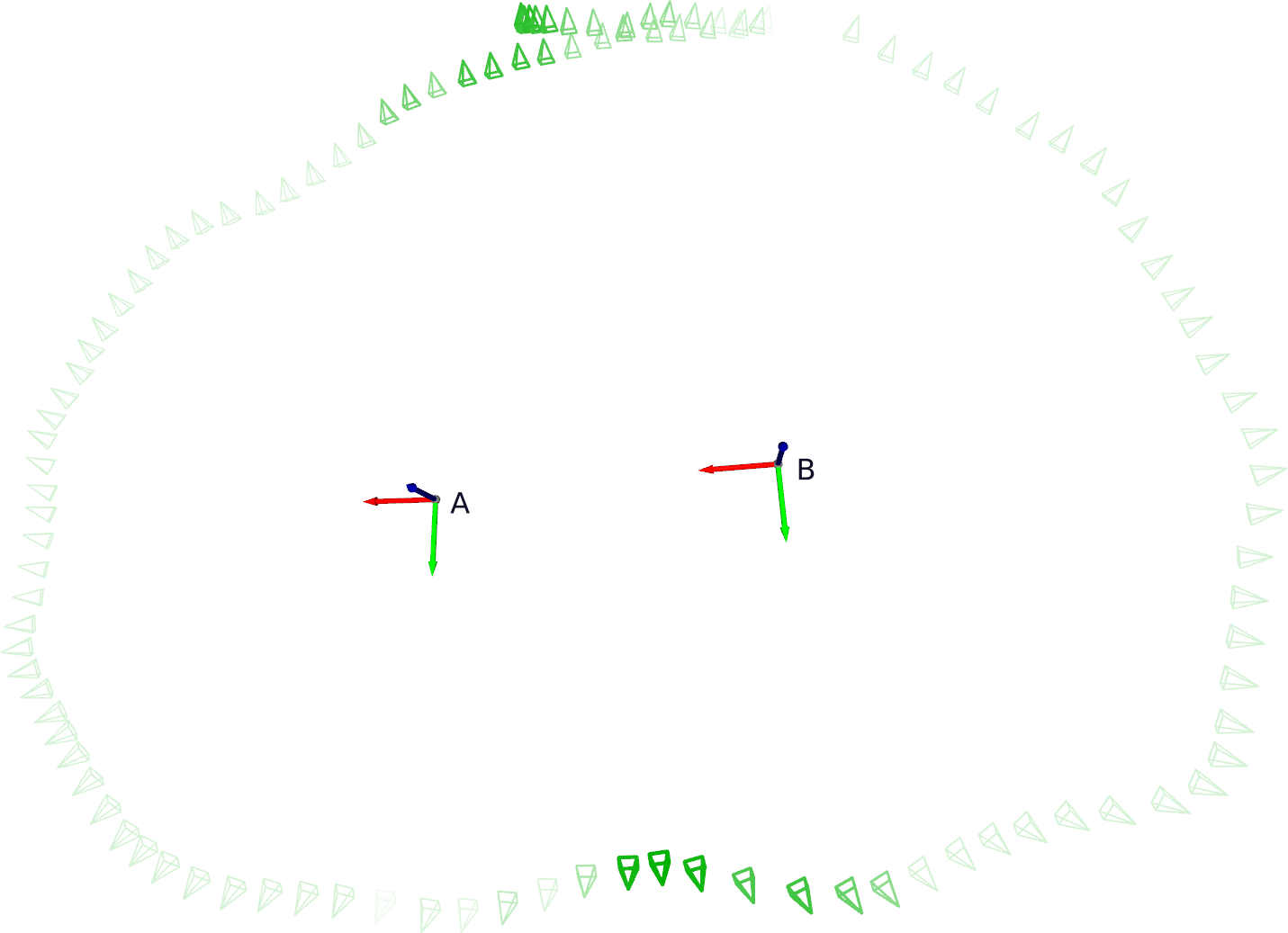}
    \caption{A visualization of how \emph{IDW-Sample} performs against \emph{Nearest}. The axes denote the reference frames for two input NeRFs, while the green camera frusta denote the test camera poses.  The darker the camera frusta, the better \emph{IDW-Sample} performs compared to \emph{Nearest}. The best-performing poses are those that evenly observe the scene.}
    \label{fig:pose-ablation}
\end{figure}

\paragraph{Ablation on Blending Performance over Query Poses}
We provide a qualitative ablation study to showcase the performance of our proposed blending method \emph{IDW-Sample} against baselines for different test poses with respect to two NeRF centers in Figure~\ref{fig:pose-ablation}. The study is supposed to provide a geometric sense of where the blending method gives the most benefits. 

\section{Conclusion}

We have introduced \Acronym, a NeRF fusion pipeline that registeres and blends arbitrary many NeRFs treated as input data. To address the problem of registration, we propose \emph{registration from re-rendering}, taking advantage of NeRF's ability to synthesize high quality novel views. To address the problem of blending, we propose \emph{IDW-Sample}, leveraging the ray sampling nature of NeRF rendering. 
While we have demonstrated \Acronym's robust performance in multiple scenarios, it inherits any of the failure cases of the input NeRFs. 
Variants that use structured priors can easily be integrated into our framework, since it is agnostic to the source NeRFs.  We believe this tool will help enable the increased proliferation of implicit representations as raw data for future 3D vision applications.

{\small
\bibliographystyle{ieee_fullname}
\bibliography{egbib}
}

\section{Appendix}

\subsection{Ablation of $\gamma$ in IDW-based Blending on the Mission Bay Dataset}\label{sec:gamma-ablation-mb}
\begin{figure}[!t]
    \centering
    \includesvg[width=\linewidth]{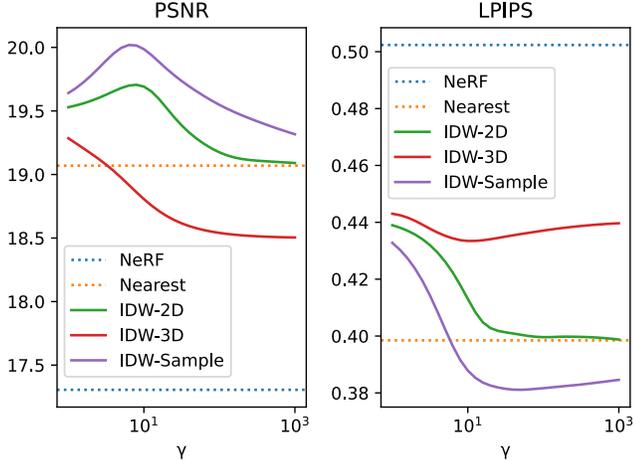}
    \caption{Effect of the blending rate $\gamma$ ranging from $[1, 10^3]$. Blending quality first increases with $\gamma$, then starts to decrease as $\gamma$ increases further. For any given $\gamma$, our \emph{IDW-Sample} works the best.}
    \label{fig:gamma-ablation-mb}
\end{figure}

We additionally present the blending rate $\gamma$ ablation on Mission Bay Dataset. Specifically, we use ground-truth transformations and set distance test ratio $\tau=1.2$. We geometrically sample $\gamma$ in $[1,10^3]$. For each sampled $\gamma$, we blend NeRFs with all IDW-based methods and report the results averaged over test images of all scenes from the Mission Bay Dataset. Since \emph{Nearest} and \emph{NeRF} are not affected by $\gamma$, we draw dotted horizontal lines for comparison. The results are shown in Figure~\ref{fig:gamma-ablation-mb}.

\subsection{Connection to KiloNeRF}\label{subsec:kilo-nerf}
In this section we establish a relationship between IDW-Sample and KiloNeRF~\cite{Reiser2021KiloNeRFSU}. Within the framework of our \emph{IDW-Sample} method, KiloNeRF is a special case when the blending rate $\gamma\rightarrow\infty$. Intuitively, KiloNeRF employs a grid of small NeRFs within the axis-aligned bounding box of the scene, where each small NeRF is only responsible for the spatial cube it occupies. Specifically, given a sample $(\delta, t)$ on ray $(\bm{o},\bm{d})$, KiloNeRF first determines which grid it falls into based on the ray sample location $\bm{o}+t\bm{d}$. The NeRF corresponding to this grid will be given a weight of $1$, while all other NeRFs will be weighted $0$. In \emph{IDW-Sample}, if we set the power $\gamma$ to infinity, it means that, for each ray sample, it will only take the field information from the closest NeRF, but not a weighted sum of information from all NeRFs. Thus, only the closest NeRF becomes responsible for that sample, which resembles the case for KiloNeRF. Our method generalizes this approach, since we can freely choose a $\gamma$ smaller than infinity to tune the range that each NeRF is responsible for, which results in better rendering quality (see how \textcolor{violet}{\emph{IDW-Sample}} in Figure~\ref{fig:gamma-ablation-mb} degrades as $\gamma\rightarrow\infty$). An ablation study of $\gamma$ is provided on both the Object-centric Indoor Scenes (in the main document) and the Mission Bay Dataset (in subsection~\ref{sec:gamma-ablation-mb}).

\end{document}